%% file: main.tex
\documentclass[acmtog]{acmart}

\usepackage{multirow} % approved by acm
\usepackage{xcolor} % approved by acm
\usepackage{xspace} %approved by acm
\usepackage{overpic}
\graphicspath{{figures/}}

\usepackage[ruled]{algorithm2e} % approved by acm
\SetAlFnt{\small}
\SetAlCapFnt{\small}
\SetAlCapNameFnt{\small}
\SetAlCapHSkip{0pt}

\setcopyright{cc}
\setcctype{by-sa}
\acmJournal{TOG}
\acmYear{2025} \acmVolume{44} \acmNumber{6} \acmArticle{202} \acmMonth{12} \acmPrice{}\acmDOI{10.1145/3763306}

\author{SaiKiran Tedla}
\orcid{0000-0002-2679-2881}
\email{tedlasai@yorku.ca}
\affiliation{%
	\institution{York University}
	\country{Canada}
}

\author{Kelly Zhu}
\orcid{0009-0002-1533-6167}
\email{zhu@cs.toronto.edu}
\affiliation{%
	\institution{University of Toronto}
	\country{Canada}
}
\affiliation{%
	\institution{Vector Institute}
	\country{Canada}
}
\author{Trevor Canham}
\orcid{0000-0003-4830-4786}
\email{tcanham@yorku.ca}
\affiliation{%
	\institution{York University}
	\country{Canada}
}
\author{Felix Taubner}
\orcid{0000-0002-9717-7181}
\email{ftaubner@cs.toronto.edu}
\affiliation{%
	\institution{University of Toronto}
	\country{Canada}
}
\affiliation{%
	\institution{Vector Institute}
	\country{Canada}
}
\author{Michael S. Brown}
\orcid{0000-0002-9840-0795}
\email{mbrown@eecs.yorku.ca}
\affiliation{%
	\institution{York University}
	\country{Canada}
}
\author{Kiriakos N. Kutulakos}
\email{kyros@cs.toronto.edu}
\orcid{0000-0002-5165-902X}
\affiliation{%
	\institution{University of Toronto}
	\country{Canada}
}
\affiliation{%
	\institution{Vector Institute}
	\country{Canada}
}
\author{David B. Lindell}
\orcid{0000-0002-6999-3958}
\email{lindell@cs.toronto.edu}
\affiliation{%
	\institution{University of Toronto}
	\country{Canada}
}
\affiliation{%
	\institution{Vector Institute}
	\country{Canada}
}


\definecolor{lightgray}{RGB}{200, 200, 200}

\input{macros}

\begin{document}

% \title{A Blurry Image is Worth a Video }
% \title{Generating Sharp Videos from Blurry Images}
% \title{Generating the Past, Present and Future from a Blurry Image}
\title{Generating the Past, Present and Future from a Motion-Blurred Image}
% \title{Recovering the Past, Present and Future from a Motion-Blurred Image}

\begin{teaserfigure}
  \includegraphics[width=\textwidth]{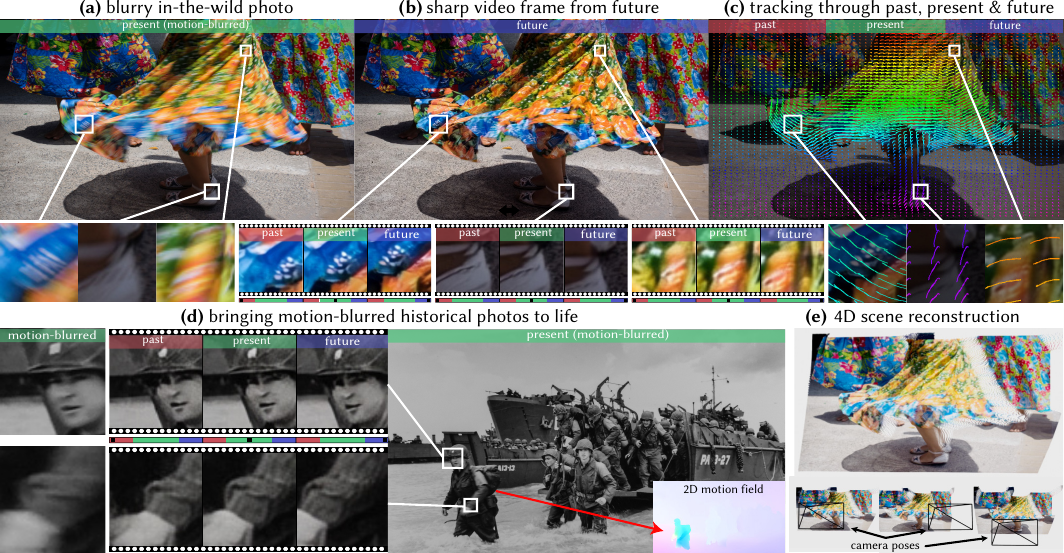}
  \caption{(\textbf{a}) Given a motion-blurred input image, our approach uses a large-scale video diffusion model to generate frames that reveal scene motion during the exposure and predict what may have occurred just before and after the image was captured. We illustrate scene motion predicted by our method with (\textbf{b}) an output video frame and (\textbf{c}) tracking from an off-the-shelf method~\cite{karaev2024cotracker}. The resulting videos capture complex scene dynamics, enabling downstream applications including (\textbf{d}) bringing historical images to life: we show insets of three sharp generated video frames (red, green, and blue bars indicate each frame's exposure window) and
  visualize subtle motions revealed by the video as a 2D motion field computed by RAFT~\cite{teed2020raft}.
%  We show the 2D motion field between the first and last generated frame  and insets of the deblurred video frames (the red, green, and blue bars indicate each frame's exposure window).
(\textbf{e}) We can also recover dynamic 3D structure and camera poses by applying a recent structure from motion technique to our output video~\cite{Li2024megasam}. \textbf{\emph{Video results are included in the supplemental webpage.}} \textit{Photos: (top) \textcopyright{} Thales Antônio, iStock; (bottom) U.S. National Archives and Records, public domain.}} 
  \label{fig:teaser}
\end{teaserfigure}

\input{sections/abstract}
\maketitle

\input{sections/introduction}
\input{sections/related_work}
\input{sections/method}
\input{sections/dataset}
\input{sections/experiments}
%\input{sections/figures}

\input{sections/assessment_generation}

\input{sections/conclusion}

\input{sections/acknowledgements}

\bibliographystyle{ACM-Reference-Format}
\bibliography{ref}

%\input{sections/figures}

% \clearpage
% \input{supp}

% \appendix

\end{document}

%% file: macros.tex
\newcommand{\img}{\ensuremath{I}\xspace}
\newcommand{\latentimg}{\ensuremath{\tilde{I}}\xspace}
\newcommand{\vid}{\ensuremath{V}\xspace}
\newcommand{\latentvid}{\ensuremath{\tilde{V}}\xspace}
\newcommand{\noise}{\boldsymbol{\epsilon}}

\newcommand{\irradiance}{\ensuremath{E}\xspace}
\newcommand{\tval}{\ensuremath{t}\xspace}
\newcommand{\exptime}{\ensuremath{\mathcal{T}}\xspace}
\newcommand{\latentexptime}{\ensuremath{\tilde{\mathcal{T}}}\xspace}
\newcommand{\tstart}{\ensuremath{t_\text{start}}\xspace}
\newcommand{\tend}{\ensuremath{t_\text{end}}\xspace}
\newcommand{\numframes}{\ensuremath{F}\xspace}
\newcommand{\numlatentframes}{\ensuremath{\tilde{F}}\xspace}
\newcommand{\cond}{\ensuremath{\mathcal{C}}\xspace}

%% file: sections/abstract.tex
\begin{abstract}
We seek to answer the question: what can a motion-blurred image reveal about a scene's past, present, and future? Although motion blur obscures image details and degrades visual quality, it also encodes information about scene and camera motion during an exposure. Previous techniques leverage this information to estimate a sharp image from an input blurry one, or to predict a sequence of video frames showing what might have occurred at the moment of image capture. However, they rely on handcrafted priors or network architectures to resolve ambiguities in this inverse problem, and do not incorporate image and video priors on large-scale datasets. As such, existing methods struggle to reproduce complex scene dynamics and do not attempt to recover what occurred before or after an image was taken. Here, we introduce a new technique that repurposes a pre-trained video diffusion model trained on internet-scale datasets to recover videos revealing complex scene dynamics during the moment of capture and what might have occurred immediately into the past or future. Our approach is robust and versatile; it outperforms previous methods for this task, generalizes to challenging in-the-wild images, and supports downstream tasks such as recovering camera trajectories, object motion, and dynamic 3D scene structure. Code and data are available at \href{https://blur2vid.github.io}{\texttt{blur2vid.github.io}}
\end{abstract}

%% file: sections/introduction.tex
\section{Introduction}

\begin{quote}
\emph{``Only photography has been able to divide human life into a series of moments, each of them has the value of a complete existence.''} \\
\hfill --- Eadweard Muybridge (attributed)
\end{quote}
A motion-blurred image is produced when the camera or scene moves during an exposure, causing scene content to smear across the image. 
Typically, motion blur is undesirable as it obscures image details and degrades visual quality, rendering an image unusable for downstream tasks.
An alternative view, however, is that motion blur 
can be highly informative about a scene's dynamics because
it encodes spatiotemporal information over the time of capture.
As such, blurred images can potentially be exploited to analyze the motions in a scene~\cite{Wang2023omnimotion,karaev2024cotracker}, to recover 3D scene information~\cite{Li2024megasam}, and to draw inferences about what occurred just before or just after a given shot~\cite{Vondrick2016cvpr}.
Inspired by recent advances in large video diffusion models~\cite{Wang2025videosurvey}---which can generate plausible videos from limited input~\cite{yang2024cogvideox}---we consider the 	question: \textit{what can a single motion-blurred image reveal about a scene's past, present, and future?}

Prior work related to this question has formulated motion blur analysis as an image restoration problem---that is, recovering a single sharp image corresponding to a specific moment within the exposure (the scene's ``present''). This is a long-standing ill-posed inverse problem, initially tackled with classical optimization techniques~\cite{Perrone2016tvdeconv} and hand-crafted deblurring priors~\cite{fergus2006removing,Levin2009blind}.
Closer to our line of inquiry, deep networks~\cite{nah2017deep} and generative models~\cite{xiao2024ddgans} have improved deblurring performance by learning a function that maps motion-blurred images to their restored counterpart. Intriguingly, by learning several such restoration functions, each tuned to a different moment within the exposure, it is now possible to map a motion-blurred photo to a short video clip~\cite{jin2018learning}.

Despite this progress in revealing a scene's present with a video, existing methods often struggle with complex scene dynamics and rapid motions. These methods train on tens of thousands of input-output pairs of blurry and sharp images, implicitly treating them as a prior on scene motion and appearance during the exposure~\cite{pham2023hypercut,zhong2023blur}. 
%However, motion blur in casually-captured photos is far too diverse to model accurately with datasets of this size,
%because its causes are many---deformations, independent motions, occlusions and disocclusions, camera shake, and the wide range of possible shutter speeds can all affect it.
However, motion blur in casually-captured photos is far too diverse to be accurately modeled with datasets of this size due to the sheer number of contributing factors, including object deformations, independent motions, occlusions and disocclusions, camera shake, and a wide range of shutter speeds.

Large video diffusion models, on the other hand, are trained on millions of video clips and billions of 
images. These models have demonstrated an
ability to generate photorealistic, temporally-consistent video sequences from as little information as a text prompt~\cite{liu2024sora}. Most significantly,  they can generate plausible reconstructions of a scene's past or future appearance given a single uncorrupted input image~\cite{lu2024vtd,brooks2024video}. Recent work has also shown that these models are highly effective at solving inverse problems in imaging and sensing~\cite{chung2023diffusion,kwon2024vision,song2023pseudoinverse,kawar2022denoising,xiao2024dreamclean,chihaoui2025diip}, 
effectively acting as general-purpose priors over the space of natural images and videos.

Here, we introduce a method that repurposes such a large, pre-trained video diffusion model~\cite{yang2024cogvideox} 
to synthesize video frames before, during, and after the exposure window of a blurry image---and use those frames for tracking and 3D reconstruction (see Figure~\ref{fig:teaser}). Our method
is specifically designed to (1) leverage large-scale pre-training, 
(2) allow precise control over the exposure start time and duration of each frame,
and (3) enable predictions of past and future as well.
Our formulation essentially treats motion blur analysis as a conditional video generation problem, not one of image restoration.

Our method is robust and versatile, generalizing to challenging in-the-wild images that include scenes of dancers, concerts, sports events, deforming cloth, moving animals, cityscapes and nature scenes---and can even 
exploit motion blur in historical photos to bring them to life as short video clips.
We show our approach achieves state-of-the-art performance when predicting the present and can capably extrapolate complex scene dynamics into the past and future.
Finally, we demonstrate our output videos reveal complex camera trajectories, intricate motions, and dynamic phenomena from just one image, and can support downstream tasks such as tracking, pose estimation, and multi-view 4D reconstruction.

%% file: sections/related_work.tex
\section{Related Work}
% Our work connects to previous techniques for blind deconvolution, recovering video from motion-blurred images, and recent methods for video generation based on diffusion models.

\paragraph{Blind deconvolution.}
Similar to our problem, blind deconvolution takes as input a single blurred observation, but seeks to explain it as the convolution of a sharp image with a spatially-invariant motion blur kernel~\cite{kundur1996deconv,fergus2006removing,shan2008high,krishnan2011blind,cho2009fast,Levin2009blind}. Spatially-varying motion blur can be restored to some extent by constraining the blur kernel to a low-dimensional manifold~\cite{hirsch2011fast}, incorporating image self-similarity~\cite{michaeli2014blind}, or using deep learning~\cite{noroozi2017motion,sun2015learning}.
Still, there are issues with using such approaches to restore ``in the wild'' motion blur,  
where camera motion and scene-dependent effects such as parallax, deformations, and occlusions--disocclusions
preclude the use of simple neural models or small datasets.

\paragraph{Video from motion-blurred images.}
% use a data driven approach
% favaro's work also implicitly assumes blur kernel because it predicts a middle image
% other work predicts all video frames directly but doesn't leverage large-scale priors
% also designs hand-crafted architectures or loss functions to account for frame ambiguity
% we leverage diffusion model to capture distribution of frame orders/outputs, much larger scale of data
Jin et al.~\shortcite{jin2018learning} introduced the problem of restoring several video frames from a blurry image.
They found that a key challenge for this task is that the restored frames can take on any order; for example, they could be played forwards, backwards, or in a shuffled order and still reproduce the blurry image when averaged together.
To resolve this ambiguity, Jin et al.\ trained a network to first restore a frame corresponding to  the middle of the exposure, and then sequentially restore adjacent frames. 
Building on this idea, subsequent works have explored a variety of solutions, incorporating carefully designed priors~\cite{li2021affine,zhang2021exposure,pham2023hypercut}, loss functions~\cite{zhang2020every,niu2021contiguous,purohit2019bringing}, and network architectures~\cite{zhong2022animation,zhong2023blur}.
%Since then, 
%several methods have been proposed to address this ambiguity problem with  specially-designed  priors~\cite{li2021affine,zhang2021exposure,pham2023hypercut}, loss functions~\cite{zhang2020every,argaw2021restoration,niu2021contiguous,purohit2019bringing}, and network architectures~\cite{zhong2022animation,zhong2023blur}. 
These approaches have trouble handling 
the blur found in casually-captured photos because their curated datasets are far too limited to be representative of real---world blur (a few thousand video clips at most).

\begin{figure}
    \includegraphics[]{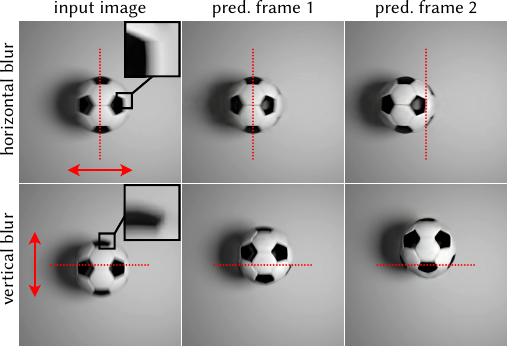}

    \caption{When given a motion-blurred image of a soccer ball as input, an off-the-shelf video diffusion model~\cite{runway2025gen4} generates a video that is consistent with the direction of motion blur. In this example, the model correctly predicts the soccer ball's horizontal or vertical motion when given the same text prompt ``a soccer ball moving without spinning.''}
    \label{fig:soccer-ball}

\end{figure}

\paragraph{Single-image animation.} Similar to our goal, single-frame animation~\cite{siarohin2019first, holynski2021animating} aims to generate video sequences given a \textit{sharp} image with no motion blur. This is accomplished by representing motion information through motion fields~\cite{holynski2021animating}, driving videos~\cite{siarohin2019first}, or motion textures~\cite{li2024generative}, and then warping and rendering the input image. Our work aims to derive motion information from the motion blur present within the input image itself while also deblurring to recover the original sharp frames.

\paragraph{Leveraging large image diffusion models.} Internet-scale image datasets \cite{byeon2022700m,schuhmann2022laion} 
have led to powerful diffusion models that can synthesize photorealistic images from a text prompt~\cite{saharia2022imagen}. 
With large diffusion models increasingly becoming available pre-trained and open-source, emerging techniques are repurposing these models as generic image priors for a variety of tasks~\cite{sun2024dreamcraft,taubner2025cap4d}. Closest in spirit to our work, Xiao et al.~\shortcite{
xiao2024dreamclean} and Chihaoui and Favaro~\shortcite{chihaoui2025diip} use pre-trained image models to generate high-quality images from degraded inputs without requiring knowledge of the specific degradation process.  While our approach also leverages a large-scale pre-trained model  in order to remain agnostic about the specific causes of motion blur (i.e., camera motion, scene motion, etc.) our use of a video model is \emph{not} agnostic to the degradation itself: motion blur is fundamentally due to time-varying appearance over the exposure, and video clips serve as a complete and physically-accurate explanation of this degradation.

\paragraph{Leveraging video diffusion models.} 
Concurrent with our work, Pang et al. \shortcite{pang2025image} pre-train a video diffusion model on small synthetic and captured datasets in order to generate video from a motion-blurred image of a robot arm. Their method was evaluated only on images with simulated blur and---as in past research on recovering videos from motion-blurred images---does not leverage large-scale pre-trained models as we do.

\paragraph{What do large video diffusion models know about motion blur?} 
Large-scale video models already encode a great deal of information about the relation between 
3D geometry and time-varying appearance, despite not being explicitly supervised on such a relation~\cite{li2024sora3D,brooks2024video}. Most pertinent to our setting, the size and diversity of 
their video datasets implies that motion blur of various causes and degrees is already
part of their training set. We thus posit that 
these models already have strong intrinsic priors over the input to our 
method---a motion-blurred image---and its connection to scene dynamics. Figure~\ref{fig:soccer-ball} shows a preliminary experiment that is highly suggestive of such priors: when given a text prompt and a 
motion-blurred image as conditioning input, a recent large-scale video diffusion model~\cite{runway2025gen4}
predicts future video frames that are consistent with the image's motion blur.
%We seek to build on this capability and fine-tune large-scale video diffusion models to predict video frames before, during, and after the moment of capture, with precise control over the exposure start time and duration of each frame.
We aim to build on this capability by fine-tuning large-scale video diffusion models to predict frames occurring before, during, and after the moment of capture, with precise control over each frame’s exposure start time and duration.

%For example, in Figure~\ref{fig:soccer-ball} we show the output of a recent large-scale video diffusion model~\cite{runway2025gen4} conditioned on horizontally and vertically motion-blurred images of a soccer ball.
%The model correctly infers future video frames that are consistent with the dynamics implied by the direction of the motion blur. 
%We seek to build on this capability and fine-tune large-scale video diffusion models to predict video frames before, during, and after the moment of capture, with precise control over the exposure start time and duration of each frame. \david{add what text prompt we used. ``A soccer ball moving without spinning.''}

%% file: sections/method.tex
\section{Motion Blur \& Video Modeling}
\label{sec:blur-modeling}
% image I^{(p)}
% pixel p
% irradiance E(t)
% time t 
% time interval 

We assume a general model for motion blur and use a large, pre-trained video diffusion model to represent the space of natural videos. 
Under this model, an image $\img$ captures the time-varying irradiance $\irradiance(\tval)$ at the sensor plane as
\begin{equation}
    I_\exptime = g\left(\int_{t\in\exptime} E(t) \, \mathrm{d}t\right),
    \label{eq:motion-blur}
\end{equation}
where $g$ is the camera response function~\cite{debevec1997recovering}, which maps the integrated irradiance to the measured image intensity, and $\exptime = [\tstart, \tend]$ is the exposure time interval.
If the irradiance at the sensor plane changes over time due to camera or scene motion, then Equation~\ref{eq:motion-blur} results in a blurry image.
Given such a motion-blurred image, we seek to recover a video \mbox{$\vid=[I_{\exptime_1}, \ldots, I_{\exptime_\numframes}]$} consisting of $\numframes$ images with sequential, and possibly non-consecutive, exposure intervals $[\exptime_{1}\ldots\exptime_{\numframes}]$.

\paragraph{Video diffusion model.}
Video diffusion models learn the probability of a video \vid, optionally conditioned on signals \cond consisting of text, image, or video data~\cite{ho2022video,xing2024survey}:
\begin{equation}
    P(\vid \, | \, \cond).
\end{equation}
Sampling from this distribution is accomplished by initializing the output video with standard Gaussian noise and then using the video model to iteratively denoise it,
through a reverse diffusion process~\cite{ho2020denoising}. 
To improve computational and memory efficiency, most video models represent their output in the compressed latent space of a pre-trained video encoder~\cite{blattmann2023align,blattmann2023stable}.
This enables the model to predict a low-resolution latent video \latentvid, where each latent frame encodes multiple high-resolution frames of the output video.
Once the reverse diffusion process completes, a pre-trained video decoder network
recovers the high-resolution output \vid from the latent video \latentvid.

\begin{comment}

\paragraph{Video diffusion model.}
We represent the distribution of natural videos using a pre-trained video diffusion model.  
These models learn the probability of a natural video \vid, optionally conditioned on signals \cond, which might consist of text, image, or video information~\cite{ho2022video,xing2024survey}:
\begin{equation}
    P(\vid \, | \, \cond).
\end{equation}
Sampling from this distribution is accomplished by initializing the output video with standard Gaussian noise and then using the video diffusion model to iteratively denoise the output video following a reverse diffusion process~\cite{ho2020denoising}. 
To improve computational and memory efficiency, most video diffusion models represent the output video in the compressive latent space of a pre-trained video encoder~\cite{blattmann2023align,blattmann2023stable}.
In this way, the video model can predict a low-resolution latent video \latentvid, where each latent frame encodes multiple high-resolution output video frames.
After the reverse diffusion process is completed, the output video \vid is recovered from the predicted latent video \latentvid using the pre-trained video decoder network.
\end{comment}

\begin{figure*}[t]
\begin{overpic}[width=\textwidth]{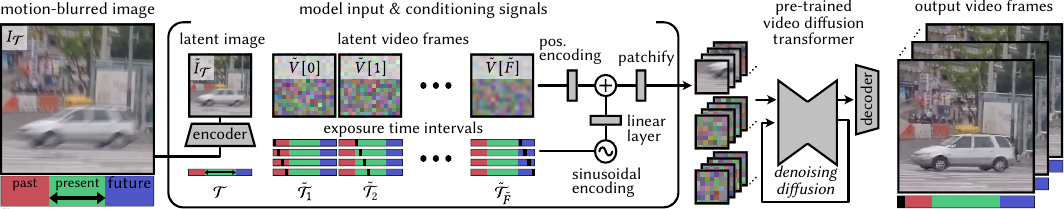}

\end{overpic} 

  \caption{Model overview. Our approach takes as input a motion-blurred image $\img_{\exptime}$. We encode it into the latent space of a pre-trained video diffusion transformer and concatenate the resulting latent image $\latentimg_{\exptime}$ with a set of noisy latent video frames $\latentvid$. 
  The latent video frames are associated with exposure times $\latentexptime_{1}, \ldots, \latentexptime_{\numlatentframes}$, which control the exposures of the video frames we seek to generate, i.e., in the past, present, or future relative to the motion-blurred image. Each latent video frame encodes multiple output video frames and so is associated with multiple exposure times. We modulate the latent frames using position encoding~\cite{su2024roformer} to incorporate information about each latent pixel's spatial position and frame index. The exposure times are encoded using a sinusoidal encoding and linear projection, and we add the result to the latent frames. The resulting latent frames are patchified, passed as input to the video model, and the model generates denoised latent frames that are decoded to recover the output video. \textit{Photos from the GoPro dataset~\cite{nah2017deep}}}.
  \label{fig:model}
      % Our approach takes as input a motion-blurred image $\img_\exptime$ a set of exposure time intervals $[\latentexptime_1, \ldots, \latentexptime_F]$ and aims to recover predicted video frames corresponding to the input exposures. 
      % We encode $\img$ into a latent image $\latentimg$, and the exposure time intervals are processed with a sinusoidal encoding, concatenated together, and passed through a linear layer before being added to the noisy latent video frames $\latentvid$. Finally, $\latentimg$ and $\latentvid$ are patchified, and $\latentvid$ is denoised by the diffusion transformer and decoded into the predicted video frames.}

\end{figure*}

\section{Generating the Past, Present and Future}

We modify a pre-trained video diffusion model to enable conditioning on both (1) a
motion-blurred image and (2) a set of exposure intervals corresponding to the individual frames of the output video:
\begin{equation}\label{eq:latent}
    P(\latentvid \, | \, \latentimg,\, \exptime_{1}\ldots\exptime_{\numframes}),
\end{equation}
where $\numframes$ is the number of generated frames  and
both the output video and the motion-blurred image are encoded in the latent space.
This conditioning gives considerable flexibility in the timespan of videos generated by the model. For instance, we can output videos that 
predict the ``present'' by choosing 
$\exptime_{1}\ldots\exptime_{\numframes}$ to all lie within the input image's actual exposure interval.
Alternatively, by conditioning the model on intervals that lie outside the actual exposure, 
we can output videos that extend 
into the ``past'' and/or the ``future.''

\subsection{Architecture}
\label{sec:architecture}
\sloppy{
Our approach builds on the text-to-video model CogVideoX-2B~\cite{yang2024cogvideox}.
This is a 2-billion parameter diffusion transformer~\cite{peebles2023scalable} that we modify to support the conditioning of Equation~\ref{eq:latent}.
Figure~\ref{fig:model} shows an overview of the model.
}

%based on an input motion-blurred image and a sequence of exposure time intervals associated with each generated video frame. 

\paragraph{Latent space encoding.} We encode the motion-blurred image into a single latent frame using the pre-trained variational auto-encoder (VAE) associated with the CogVideoX model. 
The VAE compresses the spatial resolution of the input blurry image by a factor of eight, and for videos it also compresses the temporal resolution by a factor of four. 
% Hence, to produce $\numframes$ output video frames requires an additional $\numframes/4$ latent frames to be passed as input to the diffusion transformer.
Hence, the model processes a latent video $\latentvid$ with dimensions $(\numlatentframes, \tilde{D}, \tilde{H}, \tilde{W})$, where $\numlatentframes = \numframes / 4$ is the number of latent frames, $\tilde{D}$ is the dimension of the latent space, and $\tilde{H}$ and $\tilde{W}$ are the height and width of the latent frames.
% Based on the temporal compression ratio, each latent video frame is associated with four output video frames and their exposure time intervals 
Based on the temporal compression ratio, the $i$-th frame of the latent video is associated with
four exposure  intervals, which we represent as a vector $\latentexptime_i\in\mathbb{R}^8$. For the first latent frame, for example,  
 $\latentexptime_1 = [\exptime_{1}[1],\exptime_{1}[2], \ldots, \exptime_{4}[1],\exptime_{4}[2]]$.

%associated with the $i$-th latent video frame as a vector

%as a vector $[\latentexptime_1, \ldots, \latentexptime_{\numlatentframes}]$ where, e.g.,  $\latentexptime_1 = [\exptime_{1}, \ldots \exptime_{4}]\in\mathbb{R}^8$.
% Based on the temporal compression ratio, there are four output video frames and exposure time intervals associated with every latent frame: 

%---i.e., there are $\numframes/4$ latent frames, $D$ is the dimension of the latent space, and $H/8$ and $W/8$ are the height and width of the latent frames. 

\paragraph{Position encoding.} 
Before passing the latent motion-blurred image $\latentimg$ and noise-initialized latent video frames $\latentvid$ as input to the model, we concatenate them along the temporal dimension as $[\latentimg, \latentvid[1], \ldots, \latentvid[\numlatentframes]]$. We then 
modulate them with a position encoding signal that indicates the spatial and temporal location of each latent pixel.
We follow the standard sinusoidal position encoding~\cite{su2024roformer}, adopted by CogVideoX-2B  (see Supp.\ Section~S2.1).

%, which modulates each latent pixel based on its spatial and temporal position 

\paragraph{Exposure interval encoding.} 
We also apply an exposure interval encoding to encode the start 
and end times of each latent frame.
To compute this encoding, we first assign the normalized interval $\mathcal{T} = [-0.5, 0.5]$ to the exposure of the input motion-blurred image.
Intervals $\exptime_1, \ldots \exptime_\numframes$ are then expressed
 \textit{relative} to this normalized interval.
We encode the four exposure intervals associated with a latent video frame by applying a sinusoidal positional 
encoding function $\gamma$ to each coordinate of $\latentexptime_i$, concatenating the resulting vectors,
and projecting via a linear layer:

\begin{equation}
    \textsc{Linear} \circ \textsc{Concat}\left(\gamma(\latentexptime_i[1])\ldots\gamma(\latentexptime_i[8])\right) \in \mathbb{R}^{\tilde{D}} \ .
    \label{eq:temporal-encoding}
\end{equation}
Following Vaswani et al.~\shortcite{vaswani2017attention} and Mildenhall et al~\shortcite{mildenhall2021nerf},
we define the sinusoidal positional encoding function to be
\begin{equation}
    \gamma(t) = [\cos(2\pi \nu_1 t), \sin(2\pi \nu_1 t),  \ldots, \cos(2\pi \nu_N t), \sin(2\pi \nu_N t)],
    \label{eq:pos-encoding}
\end{equation}
where $\nu_i$ is an encoding frequency and $N$ is the number of such frequencies.
Lastly, we encode the time interval of the latent input motion-blurred image in a similar fashion, after first   
replicating $\mathcal{T}$ four times to create a vector 
\mbox{$\tilde{\mathcal{T}}=[\mathcal{T},\mathcal{T},\mathcal{T},\mathcal{T}]$}
that matches the dimension of $\latentexptime_i$. See Supp.\ Section S2.1 for
additional details.

\paragraph{Fine-tuning.}
To fine-tune the diffusion transformer, we sample a motion-blurred image and its corresponding sharp video sequence. We then encode 
these frames using the VAE; add noise $\noise$ to the latent video frames according to the diffusion process; apply position and exposure-interval encoding to the noisy latent frames;  patchify them; and feed them to the video diffusion transformer. 
We optimize all parameters of the video diffusion transformer to minimize $\mathbb{E}_{\latentvid, \noise}\lVert \hat{\vid} - \latentvid \rVert_2$,
i.e., the expected L2 difference of the denoised latent video $\hat{\vid}$  and the clean latent video $\latentvid$. 

\begin{comment}
	
\paragraph{Fine-tuning.}
To fine-tune the diffusion transformer, we sample a motion-blurred image and its corresponding sharp video sequence.
These frames are encoded using the VAE, noise $\noise$ is added to the latent video frames according to the diffusion process, and the resulting noisy latent frames---modulated by the position and exposure-time encodings---are patchified and fed into the video diffusion transformer. 
We optimize all parameters of the video diffusion transformer to minimize $\mathbb{E}_{\latentvid, \noise}\lVert \hat{\vid} - \latentvid \rVert_2$,
which is the expected L2 difference between the predicted denoised latent video $\hat{\vid}$  and the clean latent video $\latentvid$. 
\end{comment}

\subsection{Implementation Details} 
\label{sec:implementation_details}

\paragraph{Optimization and inference.} 
We fine-tune the entire model, including the diffusion transformer and the linear projection layer for the exposure interval encoding, using a batch size of 64 across 16 NVIDIA L40 GPUs for 10 days. 
Training is conducted for 20,000 iterations with an AdamW optimizer~\cite{loshchilov2019decoupled} with a learning rate of $10^{-4}$. 
We also fine-tune the model for unconditional generation, by setting $\latentimg = 0$ with the same 20\% dropout percentage as the original CogVideoX model. 
At inference time, we use 50 diffusion steps with a DDPM solver~\cite{ho2020denoising} and classifier-free guidance~\cite{ho2021classifier} with a guidance scale of 1.1, which takes $\sim$2 minutes on an NVIDIA L40 GPU.

\paragraph{Fine-tuning datasets.}
To evaluate our method against baselines and to test its performance on challenging in-the-wild scenes, we fine-tune three versions of the model.
The first two are fine-tuned on specific datasets for direct comparison with baselines, as detailed in Section~\ref{sec:present}.
To enhance generalization to complex real-world imagery, we also fine-tune a third version on a diverse compilation of high-FPS videos drawn from GoPRO~\cite{nah2017deep} (240 FPS), Adobe240~\cite{su2017deep} (240 FPS), REDS~\cite{Nah_2019_CVPR_Workshops_REDS} (120 FPS), iPhone240~\cite{shimizu2023iphone} (240 FPS), and Sports240~\cite{chen2024sportsslomo} (240 FPS).
This dataset spans a variety of cameras and in-the-wild scenes, totaling 694 video clips with an average of 
507 frames per clip.
All frames were resized to the model’s native resolution of $1280 \times 720$. 

\paragraph{Blur simulation.} We simulate blur by averaging a sequence of consecutive frames from the video data
according to Equation~\ref{eq:motion-blur}. This requires
taking two considerations into account. First, since RGB colors in source videos may be stored in 
a gamma-corrected color space, which is non-linear,
we perform all frame averaging in linear sRGB space. 
Second, naive averaging of consecutive frames from a source video 
may not give a good approximation to the blur integral if there is significant ``dead time'' between them.
This occurs, for example, when the frame exposure interval is less than timespan between frames. 
To mitigate this issue, we use a frame interpolator~\cite{zhong2024clearer} to temporally upsample
all videos to 1920 FPS prior to fine-tuning. See Supp. Section S2.2 for more details.

% To compare to baselines and test our method on challenging in-the-wild scenes, we fine-tune three different versions of the model.
% The first two versions are fine-tuned on specific datasets to facilitate baseline comparisons as described in the sequel (Section~\ref{sec:present}).
% To improve the generalization capabilities of the model for complex, in-the-wild images, we also fine-tune a model on a diverse dataset consisting of high-FPS videos from GoPRO~\cite{nah2017deep} (240 FPS), Adobe240~\cite{su2017deep} (240 FPS), REDS~\cite{Nah_2019_CVPR_Workshops_REDS} (120 FPS), iPhone240~\cite{shimizu2023iphone} (240 FPS), and Sports240~\cite{chen2024sportsslomo} datasets (240 FPS). 
% The resulting videos comprise a wide range of cameras and in-the-wild scenes. 
% Overall, the compiled dataset contains \sai{X} video clips with an average of \sai{N} frames each. 
% We resize all fine-tuned frames to $1280\times720$ pixels (the native resolution of the video model) and ran a frame interpolator~\cite{zhong2024clearer} to upsample the videos to 1920 FPS. 
% The temporal upsampling is performed to remove any gap in the blurs that occur from dead times between frames (see Supp.\ Section S1). 
% Note that we perform all averaging in linear sRGB. 

\paragraph{Video generation modes.} 
We fine-tune the model so that it can operate in any one of two modes: \emph{present-only} generation 
and \emph{past/present/future} generation.
At each fine-tuning iteration we sample a random number of frames  for each mode, drawn from a batch of 
64 randomly selected videos.
% Each training batch samples uniformly from these modes.

In the present-prediction mode, we sample two to sixteen consecutive 120~FPS or 240~FPS frames from these videos.
We simulate motion blur as described above for the frames themselves (e.g., summing sixteen 1920 FPS frames to obtain one frame exposed for 1/120 seconds), and for their combined exposure interval (which can span up to 16/120 seconds).
The model must then generate these 120 or 240 FPS video frames when given their motion-blurred sum as the conditioning signal.
To improve robustness to even larger motion blurs, we sample 32 or 48 consecutive frames at 240 FPS and sum them to simulate motion blur due to an exposure interval of 32/240 or 48/240 seconds, respectively. In this case, the model must generate 16 frames that span this interval, each having an exposure of 
1/240 seconds. This allows the model to generate videos whose frames have both short exposures and 
controllable dead time (please see Section~\ref{sec:exposure_interval_control} for an analysis of this capability). 

In the past/present/future mode, we sample two to eight consecutive frames at 120 or 240 FPS, to simulate motion blur from exposures between 2/240 and 8/120 seconds.
The model must then generate four to sixteen frames that span twice that exposure, thereby predicting dynamic appearance both before and after the exposure.

%% file: sections/experiments.tex
\section{Evaluation}
% We first perform evaluation of our model.
In Section~\ref{sec:present} we compare our model to baselines on the task of generating frames during the present (during the exposure), and in Section~\ref{sec:past_future} we assess generating the past, present and the future (from before to after the exposure).

\subsection{Generating the Present}
\label{sec:present}
% For comparison with prior methods, we use the scenario of generating frames within the interval of the motion-blurred image. 
% We find that our model outperforms baselines in both scenarios.

\paragraph{Datasets.}
We first compare our model on the GoPro~\cite{nah2017deep} dataset, which contains over 3000 blurred images along with paired 7-frame videos that were used to synthetically generate the blurred images. For this comparison, our model is only trained with $1280\times720$ sequences from the GoPro~\cite{nah2017deep} dataset. To be compatible with all baselines, we downsample the model outputs to $640 \times 360$ resolution for evaluation. We also compare our model on the B-AIST++ dataset following Zhong et al.~\shortcite{zhong2022animation}. 
The dataset contains videos of dancers and corresponding bounding boxes ($192\times 160$) used to crop and assess the model output. We utilize the original train and test splits for both datasets.

\paragraph{Baselines.} 
We compare our model against the most relevant and recent baselines with publicly available code.
On the GoPro dataset, we evaluate against MotionETR~\cite{zhang2021exposure}, which reconstructs a video from a single motion-blurred image by first predicting a sharp image and an associated motion field, then warping the sharp image using the estimated motion.
We also compare to the method of Jin et al.~\shortcite{jin2018learning}, which employs an ordering-invariant loss to train networks that recover a sequence of sharp frames from a blurred input.
For the B-AIST++ dataset, we compare to Animation from Blur~\cite{zhong2022animation}, which uses coarse optical flow estimation to guide video reconstruction via a generative adversarial network~\cite{sohn2015learning}.
% We note that we contacted other baselines~\cite{purohit2019bringing, zhang2020every}, but were unable to receive code or video results. 

\paragraph{Metrics.}
Quantitative evaluation requires care due to the inherent motion ambiguities in recovering videos from a single motion-blurred image.
For instance, both a video and its time-reversed counterpart may be equally valid predictions given the same input.
To address this, prior work compares both the predicted video and its time-reversed version to the ground truth, reporting image quality metrics for the direction that yields better performance~\cite{zhong2022animation}.
While this approach accounts for global motion ambiguity at the frame level, it overlooks local motion variations across the image.
To better capture such spatially varying ambiguities, we additionally report metrics computed on forward and time-reversed \textit{patches} of the predicted videos, allowing for more fine-grained evaluation.

More formally, let $\hat{\vid}^{(p)}\in\mathbb{R}^{F \times H_p\times W_p}$ be the $p$th patch of a predicted video with $F$ frames, let $\hat{\vid}^{(p)}_\text{rev}$ be the patch at the same spatial location in the temporally-reversed video, and let $\vid^{(p)}$ be the corresponding patch in the ground truth video.
Then, we can define the bidirectional patch-based version $M_p$ of a standard image quality metric $M(\cdot, \cdot)$ as
\begin{equation}
M_p(\hat{\vid}, \vid) = \frac{1}{p}\sum_p \min\left[ M\left(\hat{\vid}^{(p)}, \vid^{(p)}\right), M\left(\hat{\vid}_\text{rev}^{(p)}, \vid^{(p)}\right) \right].
\label{eqn:bidirectional-metric}
\end{equation}
%For image metrics, such as SSIM, we compute $M$ by averaging across frames.

% More formally, for a given image quality metric $M$ with a predicted video $\hat{\vid}$, the temporally reversed predicted video $\hat{\vid}_\text{rev}$, and a ground truth video $V$, we define the metric at a given patch-size $s\times s$ as
% \begin{equation}
% M_{s \times s} = \frac{1}{F H W} \sum_{f=1}^{F} \sum_{i=1}^{H} \sum_{j=1}^{W} M_{i,j,f},
% \end{equation}
% where
% \begin{equation}
% M_{i,j,f} = \min\left( M\left(\ V_p[i,j,f], V_{gt}[i,j,f]\right),\ M\left(\ V_{rp}[i,j,f], V_{gt}[i,j,f]\right) \right),
% \end{equation}
% and $H$ and $W$ describe the number of patches in the height and width dimensions, $F$ is the number of frames, and $V_p[i,j,f]$ denotes a patch from the predicted video $V_p$.

We employ five evaluation metrics in total.
The first three are standard image-based metrics: peak signal-to-noise ratio (PSNR), structured similarity index measure (SSIM)~\cite{wang2004image}, and LPIPS~\cite{zhang2018unreasonable}.
For the GoPro dataset, we use a patch size of $1 \times 1$ for PSNR and $40 \times 40$ for SSIM and LPIPS.
For the B-AIST++ dataset, which requires evaluation on smaller crops of $192\times 160$ given by labeled bounding boxes, we use patch sizes of $ 1 \times 1$ for PSNR and $ 32 \times 32$ for SSIM and LPIPS.
When computing PSNR, we slightly modify Equation~\ref{eqn:bidirectional-metric} to better capture frame-level signal-to-noise characteristics. Specifically, for each patch, we first select either the forward or reverse MSE based on Equation~\ref{eqn:bidirectional-metric}; then, we average the patch-wise MSE to obtain a single frame-level MSE. PSNR is computed from this value, and the final score is reported as the average PSNR across all frames.
% When computing PSNR, we slightly modify Equation~\ref{eqn:bidirectional-metric} to better reflect frame-level signal-to-noise characteristics.
% We first average the patch-level mean squared error (MSE) across each frame---using either the forward or reverse mean squared error as determined by Equation ~\ref{eqn:bidirectional-metric}, then we compute PSNR from the resulting frame-level MSE, and finally report the average PSNR across all frames.
% \trevor{We average the MSE for each patch across all frames in reverse and forward directions separately (assuming that the direction of each patch should be consistent throughout the sequence in the short time intervals tested here i.e., avoiding the physically unlikely solution where the minimum error is obtained with some combination of forward and reverse direction frames). Then these frame-averaged patch-wise errors are plugged into equation 6 to find the direction with minimum error per patch. Then, we average across patches and take the PSNR.} 

For the remaining metrics, we use Fréchet Video Distance (FVD)~\cite{unterthiner2019fvd} to measure the distributional similarity between generated and ground-truth videos.
FVD is computed at full resolution using videos played in the forward direction.
Finally, we report end-point error (EPE)~\cite{zhang2020every} using the RAFT~\cite{teed2020raft} optical flow estimator.
EPE measures the difference in optical flow between the predicted and ground-truth videos, from the first to the last frame.
This metric is computed bidirectionally, selecting per pixel the temporal direction (forward or backward) that minimizes the flow error. 

\begin{figure*}[]
    \includegraphics[width=0.96\textwidth]{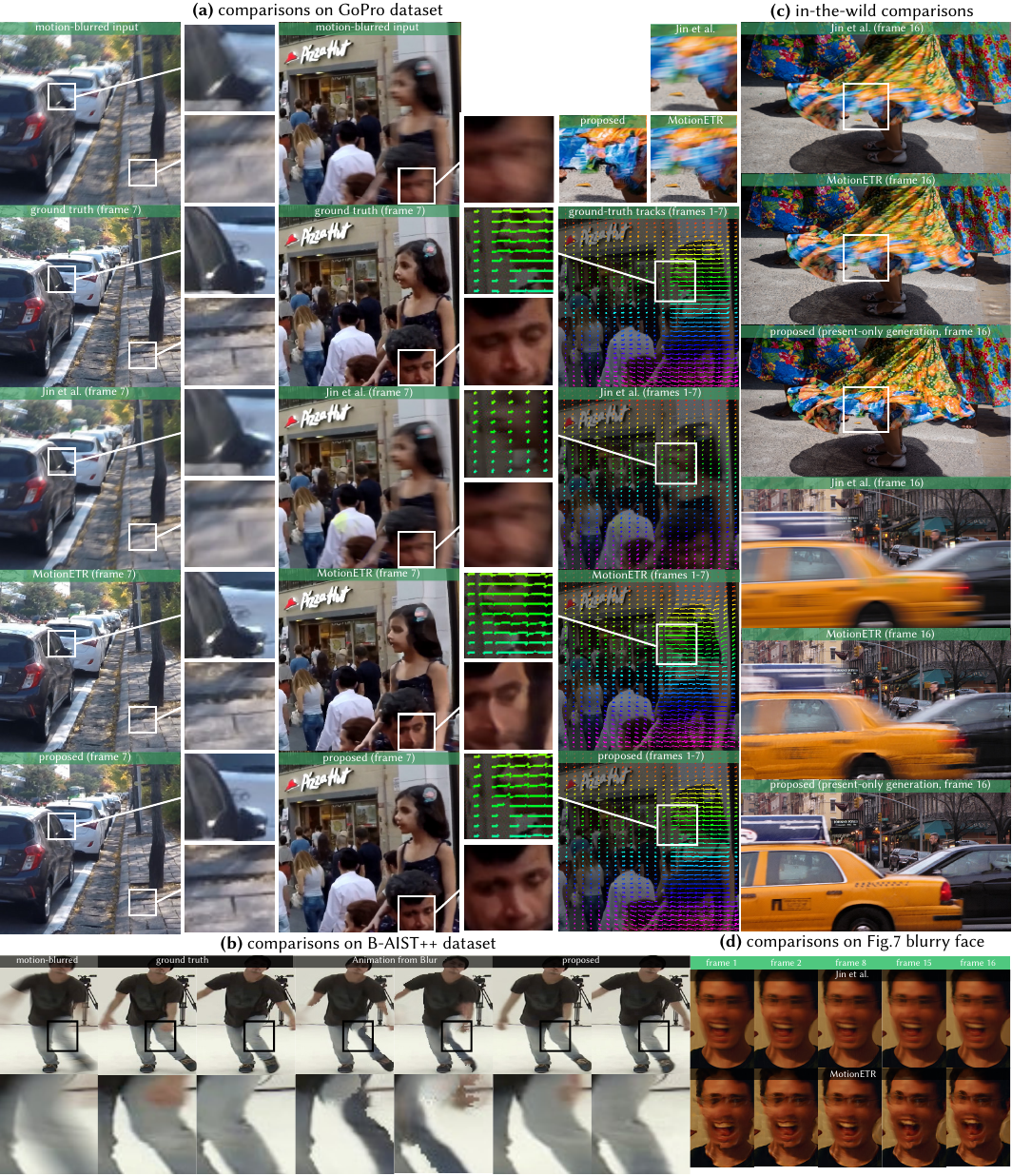}
    %\vspace{-1em}
    \caption{\textbf{(a--b)} Results on the GoPro and B-AIST++ datasets~\cite{nah2017deep,zhong2022animation}. We compare our method to MotionETR~\cite{zhang2021exposure}, Jin et al.~\shortcite{jin2018learning}, and Animation from Blur~\cite{zhong2022animation} and find that our method recovers significantly clearer output video frames with motion tracks that are more consistent with the ground-truth video sequence (tracks estimated using Cotracker~\cite{karaev2024cotracker}). \textbf{(c--d}) Additional baseline comparisons to in-the-wild data. We find that Jin et al.~\shortcite{jin2018learning} and MotionETR fail to recover sharp video frames on these challenging, in-the-wild sequences, likely due to the more limited scale of their training datasets and learned motion priors (compare to our results on these scenes in Figures~\ref{fig:teaser} and~\ref{fig:applications}).}
    \label{fig:gopro-qualitative}
\end{figure*}

\paragraph{Results} 
Across all evaluation metrics, our model consistently outperforms the baselines (see Tables~\ref{tab:present} and~\ref{tab:baist++}), demonstrating both accurate motion prediction and high image quality.
Qualitatively, our method performs well on the GoPro and B-AIST++ datasets, as shown in Figure~\ref{fig:gopro-qualitative}. Additional video comparisons are available on the supplemental webpage.

Notably, the model successfully handles challenging and complex effects, such as spatially varying blur, occlusions, and disocclusions.
Compared to baselines like MotionETR and Animation from Blur, our results appear more natural and free of warping artifacts.
We attribute this partly to the large-scale pre-training of the base video model and the resulting strong priors that it learns on natural videos ~\cite{shao2024learning}.
Overall, our generated videos not only look more realistic but are closer to the distribution of ground truth videos in the datasets as measured by FVD.

% We find across all these metrics our model outperforms baselines quantitatively across all metrics (see Tables~\ref{tab:present} and Tables~\ref{tab:baist++}).
% This shows our method is accurate in terms of motion prediction and also outperforms others in image quality.
% Additionally, our method performs well qualitatively on the GoPro dataset as seen in Figure~\ref{fig:gopro-qualitative}.
% We include comparisons with B-AIST++ and more comparisons with GoPro in our supplementary webpage.
% We notice our model smoothly transitions at occlusion boundaries as it is able to inpaint content that is not visible in the initial image.
% Next, we find our methods results to look more natural and not cause strong warping artifacts like the ones created by MotionETR and Animation from Blur.
% We believe the model is able to do this due to the geometric priors~\cite{shao2024learning} that can be found in video diffusion models.
% Finally, our videos look much more realistic as can be seen visually and also through our FVD performance.

% \paragraph{Increasing Motion Blur}
% Duis aute irure dolor in reprehenderit in voluptate velit esse cillum dolore eu fugiat nulla pariatur. Excepteur sint occaecat cupidatat non proident, sunt in culpa qui officia deserunt mollit anim id est laborum.

% \begin{figure}[h]
%     \includegraphics[width=\columnwidth]{figures/results-increasing-blur.pdf}
%     \caption{Baist qualitative}
%     \label{fig:baist-qualitative}
% \end{figure}

\subsection{Generating the Past and the Future}
\label{sec:past_future}
Our method robustly reconstructs both the timespan of capture and  subsequent motion, as demonstrated through an additional evaluation on the GoPro dataset. We omit baseline comparisons in this setting, as the baselines were not trained for this task.

% We find our method to have pleasing results across a variety of scenarios and showcase these videos in Section~\ref{sec:in-the-wild} and our supplementary webpage.
% Other baselines presented in Section~\ref{sec:present} are not designed for prediction into the past or the future. Thus, we are unable to compare with them. Instead, we seek to answer another question. \textit{How much does motion blur constrain the future, present, and the past?} 

\begin{figure}[t]
    \includegraphics[width=0.48\textwidth]{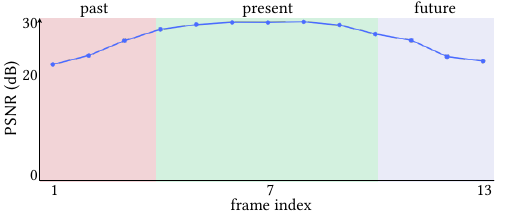}
    \caption{PSNR per frame (using best ordering of frames) when predicting 13 frames from a blurry image synthesized from 7 frames in the GoPro dataset.}
    \label{fig:past-present-future}
\end{figure}

We evaluate our model on the GoPro dataset by synthesizing blurred images from 7 sharp 240 FPS frames.
The model then predicts a 13-frame sequence: three frames before the moment of capture, seven frames during the exposure, and three frames after the capture.
We plot the bidirectional patch PSNR in Figure~\ref{fig:past-present-future}.
Interestingly, the quality remains consistent across the seven middle frames, suggesting that each frame within the motion blur is equally well constrained. 
% Deblurring is equally challenging regardless of the exposure position.
Although accuracy degrades for predictions outside the moment of capture, the video frames still fall within 20--30 dB PSNR. 
The results suggest that the motion-blurred images provide strong cues for past and future prediction.
We show qualitative examples of past and future generation in the supplemental webpage.

\begin{table}[t]
    \caption{Quantitative results on the GoPro Dataset. Note that we use the bidirectional patch versions of PSNR, SSIM, and LPIPS.}
    \small
    \resizebox{\columnwidth}{!}{%
    \begin{tabular}{lccccc}
        \toprule
        Method & PSNR$_p$ $\uparrow$ & SSIM$_p$ $\uparrow$ & LPIPS$_p$ $\downarrow$ & FVD $\downarrow$ & EPE $\downarrow$ \\
        \midrule
        Jin et al.~\shortcite{jin2018learning}       & 25.23 & 0.8190 & 0.084 & 235.53 & 3.38 \\
        MotionETR~\shortcite{zhang2021exposure}        & 26.54 & 0.8825 & 0.015 & 94.90  & 1.46 \\
        proposed   & \textbf{30.01} & \textbf{0.9359} & \textbf{0.010} & \textbf{21.46} & \textbf{0.39} \\
        \bottomrule
    \end{tabular}}
    \label{tab:present}
\end{table}

\begin{table}[t]
    \caption{Quantitative results on the B-AIST++ dataset. Note that we report the bidirectional patch versions of PSNR, SSIM, and LPIPS.}
    \small
    \resizebox{\columnwidth}{!}{%
    \begin{tabular}{lccccc}
        \toprule
               & PSNR$_p$ $\uparrow$ & SSIM$_p$ $\uparrow$ & LPIPS$_p$ $\downarrow$ & FVD $\downarrow$ & EPE $\downarrow$ \\
        \midrule
        Animation from Blur    & 26.69 & 0.9209 & 0.042 & 138.27 & 2.65 \\
        proposed   & \textbf{27.37} & \textbf{0.9306} & \textbf{0.027} & \textbf{37.16} & \textbf{1.78} \\
        \bottomrule
    \end{tabular}}
    \label{tab:baist++}
\end{table}


% This robustness is also evident in the qualitative results available on our supplementary webpage.

% We use the GoPro dataset and synthesize blurred images from 7 sharp frames.
% Then, we use our model to predict a 13 frame sequence where 3 frames are in the past (before motion blur), 7 in the present (within the motion blur), and 3 in the future (after motion blur).
% We plot the PSNR per frame in Figure~\ref{fig:past-present-future} (choosing the best order of frames).
% Naturally, we find that the past and future predicted frames are worse than the frames within the present.
% But interestingly, we notice within the motion-blur interval the prediction quality is quite similar for all 7 frames.
% This shows that any frame within the motion blur is constrained the same by motion blur i.e., deblurring is similar regardless of exposure interval predicted.
% Additionally, we see that the motion-blurred image provides a reasonable boundary condition when generating the past and future frames.
% Thus, our output quality is not degraded significantly even for these frames.
% This is also demonstrated qualitatively in our supplementary webpage.

\begin{figure*}[]
    \includegraphics[width=\textwidth]{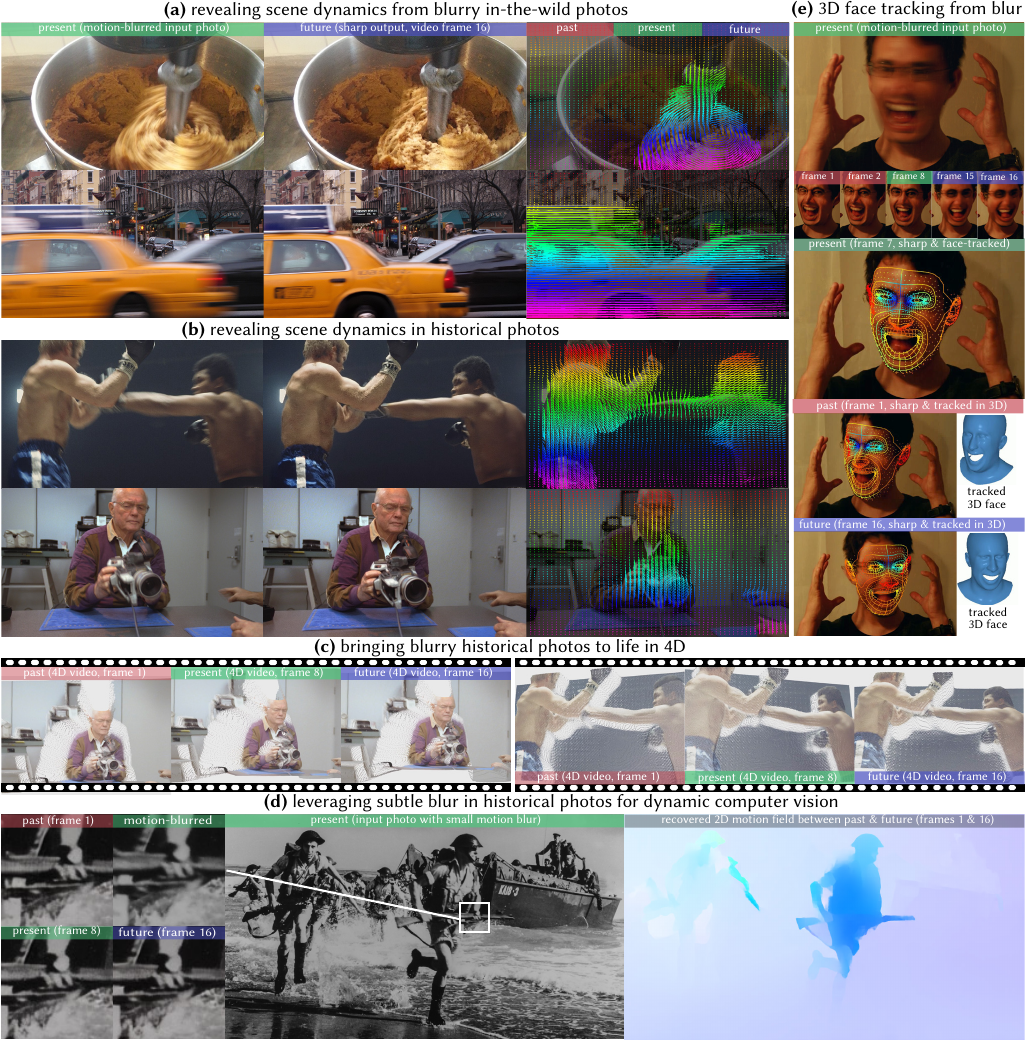}
    %\vspace{-2em}
    \caption{Applications of the proposed method (\emph{please see the video results in the supplemental webpage}). \textbf{(a)} We demonstrate generating the past, present, and future from in-the-wild motion-blurred images, as indicated by the  red, green, and blue labels. The method recovers sharp frames and scene dynamics from a mixer (top) and a busy city street (bottom). Note the complex motion trajectories recovered by applying an off-the-shelf tracker~\cite{karaev2024cotracker} to our generated videos. \textbf{(b)} By exploiting motion blur in historical photos, we reveal scene dynamics, e.g., the movement of Mohammad Ali in a boxing match or astronaut John Glenn picking up a camera. \textbf{(c)} We bring images ``to life'' by predicting 3D scene dynamics and camera poses from our generated video frames with off-the-shelf structure from motion methods~\cite{Li2024megasam}. \textbf{(d)} We even recover subtle motions in black and white photographs captured during World War II over 80 years ago. We reveal motions in the generated deblurred frames (insets) by applying an optical flow method~\cite{teed2020raft} to a past and future predicted frame. (\textbf{e}) Finally, we recover 3D facial dynamics from a motion-blurred image by applying a face tracker to our output video~\cite{taubner20243d}.  \textit{Historical photos: (1) Coast Guard Lands the British Marines (1944), U.S. National Archives and Records, public domain; (2) Senator John Glenn (1998), NASA, public domain; (3)  Mohammad Ali boxing Jürgen Blin (1971), \textcopyright{} Wikimedia Commons, CC BY-SA 4.0. (4) Blurry face, \textcopyright{} DieselDemon, Flickr, CC BY 2.0.}} 
    \label{fig:applications}
\end{figure*}

\section{Applications}
\label{sec:applications}

We test our model on a wide range of applications, including generating video from challenging in-the-wild motion-blurred photos (Section~\ref{sec:in-the-wild}), bringing historical  
motion-blurred photos to life (Section~\ref{sec:historical}), reconstructing 3D scene dynamics (Section~\ref{sec:3d}), and recovering 3D human head pose from motion-blurred portrait photos (Section~\ref{sec:human}).
% Finally, in Section~\ref{sec:human}, we show reconstruction of human and face poses. 

% \begin{table}
% \begin{tabular}{lccc}
% Metric & Favaro & MotionETR & Ours \\
% \hline
% %FID $\downarrow$ & 40.5 & \textbf{6.66} & 12.1 \\ - We can drop FID
% FVD $\downarrow$ & 237 & 106 & \textbf{30.2} \\
% LPIPS $\downarrow$ & 0.2421 & 0.1024 & \textbf{0.0804} \\
% DISTS $\downarrow$ & 0.1327 & \textbf{0.0658} & 0.0681 \\
% SSIM $\uparrow$ & 0.753 & 0.707 & \textbf{0.833} \\
% DE2000 $\downarrow$ & 4.50 & 4.70 & \textbf{4.18} \\
% PSNR $\uparrow$ & 23.6 & 22.5 & \textbf{26.1} \\
% \caption{BAIST++ Results}
% \end{tabular}
% \caption{Quantitative comparison on the BAIST++ dataset.}
% \label{tab:gopro-results}
% \end{table}

\subsection{In-the-Wild Generation}
\label{sec:in-the-wild}
We evaluate our model on motion-blurred images from in-the-wild scenes and find that it consistently produces realistic videos across a wide range of scenarios.
Figures~\ref{fig:teaser} and~\ref{fig:applications}(a) and our supplemental webpage show qualitative examples.
Our model handles diverse motion types—including running, biking, surfing, gymnastics, object manipulation, water splashes, and even complex circular motion such as children riding a merry-go-round.
The scenes span a broad spectrum of subjects, from people and animals to kitchen implements, swaying tree branches, and falling confetti.
We even recover crisp videos from multiple blurred objects with different motion trajectories, such as for bustling city intersections filled with cars and pedestrians.
We show the recovered motion in Figures~\ref{fig:teaser} and~\ref{sec:applications}(a) by applying an off-the-shelf tracking algorithm~\cite{karaev2024cotracker}, and we visualize the dense-tracked motion paths in the videos on the supplemental webpage.

Qualitative comparisons with baselines, including Jin et al. and MotionETR are provided in Figure~\ref{fig:gopro-qualitative} and the supplemental webpage.
We use their publicly available models trained on the GoPro dataset.
The baselines struggle to generalize to the varied content and motion types present in our test set and are unable to generate frames beyond the motion-blurred interval.
% \sai{Note: the baselines were not trained on the full dataset—consider clarifying this if needed.}

% We test our model on motion-blurred images from in-the-wild scenes and find it to produce realistic videos in a variety of scenarios. Please see Figure~\ref{sec:applications}a  and our supplementary webpage for examples. Our model is able to handle scenes with a diverse range of motions including running, biking, surfing, gymanstics, holding objects, water splashing, and even circular motion blurs such as kids riding a merry-go-round. Our scenarios encompass a wide range of subjects such as people, animals, kitchen utensils, and trees. Finally, our model is able to handle multiple motion-blurred objects such as when both cars and people are moving in a big city. Qualitative comparisons with baselines Jin et al. and MotionETR (trained on GoPro) are shown in the supplementary webpage. We find the baselines to not generalize well to these various scenes and again they are unable to predict outside the present.~\sai{Not sure how to mention baselines here... they were not trained on the full dataset}

\subsection{Revealing Scene Dynamics in Historical Photos}
\label{sec:historical}
We find our model generalizes to historical photos---enabling recovery of video for dynamic scenes that were photographed, in some cases, over 80 years ago. For example, we leverage subtle motion blur cues to recover the direction and magnitude of the motions of American soldiers (Figure~\ref{fig:teaser}) and British marines (Figure~\ref{fig:applications}(d)) piling from Coast Guard landing barges onto the French coast on June 6, 1944 during the Allied invasion of Normandy.  
For these results, we show 2D motion fields by applying off-the-shelf optical flow prediction~\cite{teed2020raft} to our past and future-predicted video frames.
We can also watch Mohammad Ali land a blow on Jürgen Blin in our video, generated from a photo of the 1971 boxing match. 
Finally, another result from a 1998 photo captures the astronaut John Glenn carefully handling a 24 mm camera.

% we reconstruct videos of soldiers departing their boat in World War 2, Mohammed Ali throwing a punch, and even astronauts in a space station. We show results for a variety of historical photos in our supplementary webpage. These results show that our model's understanding of motion blur is not camera specific or even restricted to color images. We believe this is an exciting application of our model, showing that the motion blur priors within video diffusion models apply to images at any historical time.

\subsection{Reconstructing 4D Scenes}
\label{sec:3d}
To reconstruct dynamic 3D (i.e., 4D) scenes from our generated videos, we apply MegaSaM~\cite{Li2024megasam} and extract dense, dynamic 3D point clouds and corresponding camera poses. 
The reconstructions preserve spatial and temporal coherence, and we show visualizations of the reconstructed camera trajectories and scene geometries in Figures~\ref{fig:teaser} and~\ref{fig:applications}(c) and in the supplemental webpage.
In this fashion, our approach can be applied to in-the-wild or historical photos---bringing them to life through 4D visualization.

Our approach enables two distinct but complementary forms of 3D understanding from a single motion-blurred image.
First, in scenes exhibiting significant rigid motion---such as turning heads or fast-moving limbs---our generated video frames reveal temporally coherent disparity cues that would otherwise be lost in a single blurred frame.
This disparity information becomes accessible only because our model produces geometrically consistent video sequences rather than merely plausible frame-by-frame generations.

Second, in dynamic scenes with non-rigid structure, our synthesized videos can be lifted to coherent 4D representations that are physically plausible in space and time.
% While monocular depth models~\cite{ming2021deep} can estimate per-frame geometry, they do not enforce consistency across frames.
% In contrast, our method enables dynamic scene reconstructions with geometry that remains coherent throughout the sequence (see dynamic point cloud visualizations in the supplemental webpage).
Specifically, we show that our output videos can be lifted into a single 4D representation of dynamic scene geometry in a consistent world coordinate space.
Such reconstructions are only possible when the input video maintains 3D consistency across time and when multi-view techniques are applied to exploit cross-frame correspondences.

% Second, in dynamic scenes lacking a single rigid structure, we show that the synthesized videos are physically plausible when lifted to 4D.
% Specifically, while any video might admit depth estimates frame-by-frame from a monocular depth model~\cite{ming2021deep}, our approach can be used to produce dynamic scene reconstructions whose geometry is consistent from one frame to the next (please see the dynamic point cloud visualizations included in the supplemental webpage).
% Such geometrically consistent outputs cannot be produced frame-by-frame from a monocular depth model; rather, they require a 3D-consistent input video sequence and multi-view reconstruction techniques that leverage cross-frame correspondences.

% The videos produced by our model are more realistic due to the video diffusion prior. Thus, we qualitatively find our videos to be geometrically consistent. We notice this from how our model handles occlusion boundaries as well as how motion is handled relatively depending on object distance. Thus, to learn more about the geometric properties of these videos we run MegaSAM~\cite{Li2024megasam} on our videos. We find that we can recover detailed 3D point clouds as well as the camera poses allowing for plausible dynamic 3D scene reconstructions. Please see our supplementary website for these visualizations.

\subsection{Recovering 3D Human Pose}
\label{sec:human}
We show that our model enables reconstruction of 3D human head pose using the disparity cues in a motion-blurred portrait photograph.
Specifically, we apply our method to the photo shown in Figure~\ref{fig:applications}(e) and recover sharp video frames showing the rigid motion of the head.
We then apply a 3D human head tracker that predicts 2D facial landmarks for each frame and registers a parametric 3D human head model to the landmarks through a joint optimization procedure across all frames~\cite{taubner20243d}.
We show the tracked 2D landmarks and the recovered animated 3D head model in Figure~\ref{fig:applications}, and we show a video animation of these results in the supplemental webpage.
This demonstrates that our predicted videos provide enough geometric consistency to be explained by a single 3D representation.

%We also find our method allows us to recover dynamic poses on humans from a single blurred image. We utilize an off-the-shelf pose prediction method~\cite{taubner20243d} and run it on video sequences generated from blurry images. In the first example, we \sai{needs to be filled in depending on the example we chooose} In the second example, we can visualize the poses of a man who is screaming while shaking his head. The results are surprisingly robust and small mistakes in pose (e.g., the left arm of the surfer) are due to viewing direction issues and could potentially be resolved by more effective pose estimation models. 

% \subsection{Small Motions}
% Finally, we note that our model is able to predict small motions that are \sai{depends on example}

%% file: sections/assessment_generation.tex
\section{Assessment of Generated Videos}
We conduct additional experiments to assess the performance of the model and the impact of architectural design choices on the generated videos. Specifically, we discuss our model's ability to capture the multi-modality of generating videos from a single motion-blurred image (Section~\ref{sec:multi_modality}), maintain consistency (Section~\ref{sec:consistency}), and control the generated frames' exposure intervals (Section~\ref{sec:exposure_interval_control}).

% This section provides a high-level overview of the evaluation and results, and additional details, figures, and quantitative analyses are provided in Supp.\ Section S1.}

% \revision{We discuss additional observations and details regarding our model, highlight a few limitations of our approach, and finally propose exciting directions for future work.}

\begin{figure*}[t]
  \includegraphics[width=0.95\textwidth]{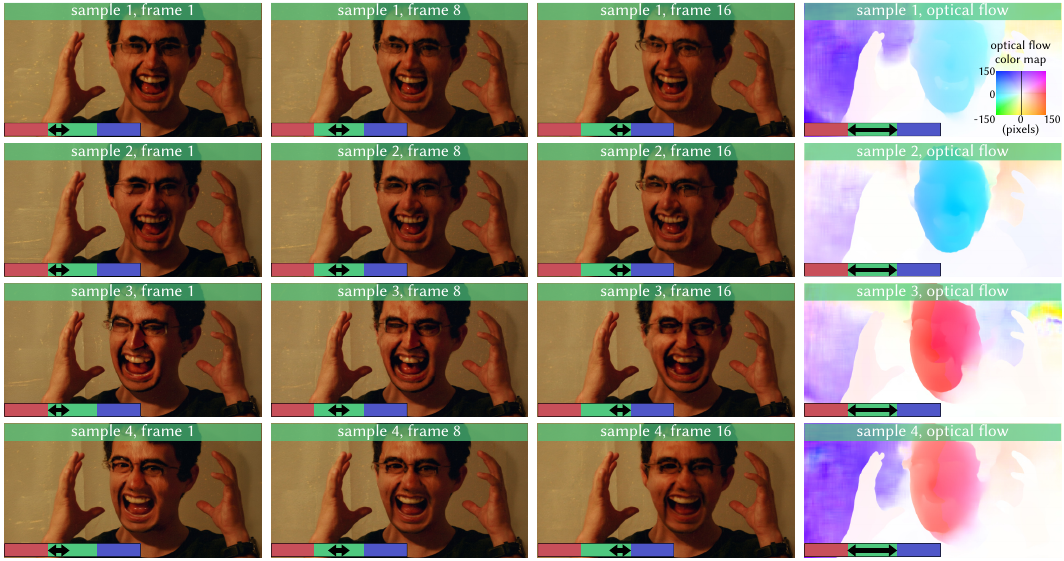}
  \caption{Four sampled videos given a motion-blurred image of a face (see Figure~\ref{fig:applications}). In the rightmost column, we show the optical flow between the first and last frame computed using RAFT~\cite{teed2020raft}. We observe that our model can generate videos where the face is moving left to right and right to left. This demonstrates the model's ability to capture the multi-modal distribution of output videos. \textit{Photo from Flickr: \textcopyright{} DieselDemon, CC BY 2.0.}}
  \label{fig:face_direction}
\end{figure*}

\begin{figure*}[h!]
  \includegraphics[width=0.95\textwidth]{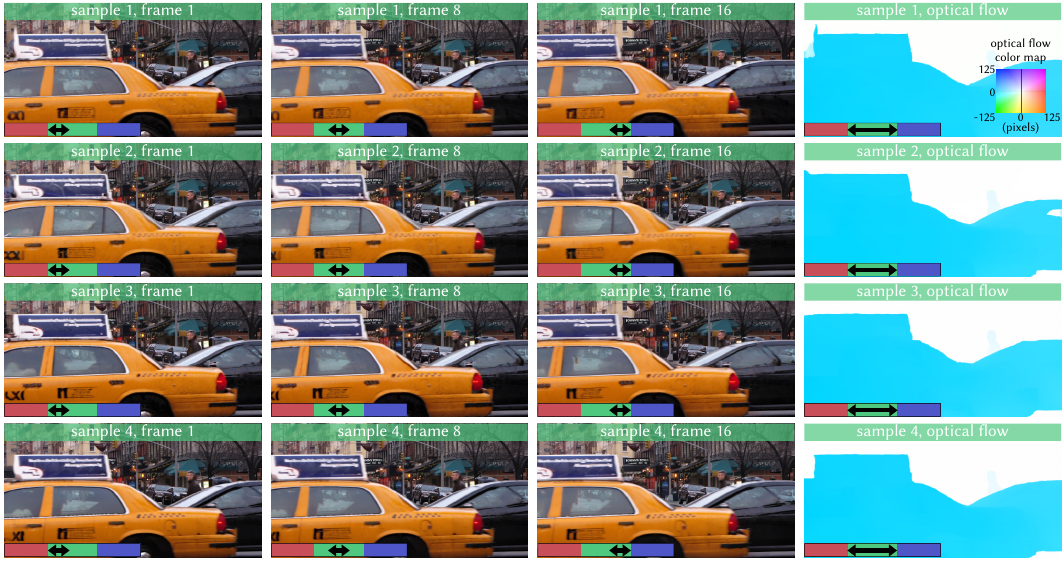}
  \caption{Four sampled videos given a motion-blurred image of a taxi (see Figure~\ref{fig:gopro-qualitative}). In the rightmost column, we show the optical flow between the first and last frame computed using RAFT~\cite{teed2020raft}. Our model consistently generates videos where the taxi moves forward (right to left), suggesting that the model incorporates real-world motion priors—for instance, the expectation that cars typically move forward on the road. \textit{Photo: \textcopyright{} Pat Ossa, CC BY 2.0}.}
  \label{fig:taxi_direction}
\end{figure*}

\subsection{Multi-Modality} 
\label{sec:multi_modality}
Motion blur in an input image could potentially be explained by an infinite number of generated videos. 
We probe the distribution of videos learned by the model by sampling and comparing multiple output videos.
Qualitatively, we find the motion in the output videos is consistent with motion behavior expected from a number of different object categories (see Figures~\ref{fig:face_direction}--\ref{fig:taxi_direction} and the supplemental webpage).
For example, humans, cars, and animals are generally predicted to be moving in the forward-facing direction.
However, in cases where motion direction is more ambiguous (e.g., a person shaking their head---see Figure~\ref{fig:face_direction}), the output videos sample multiple plausible motion directions.
Hence, the model does not \emph{exactly} recover the motion that occurred during the moment of capture but rather generates samples of what might plausibly have occurred.

\begin{figure*}[]
  \includegraphics[width=\textwidth]{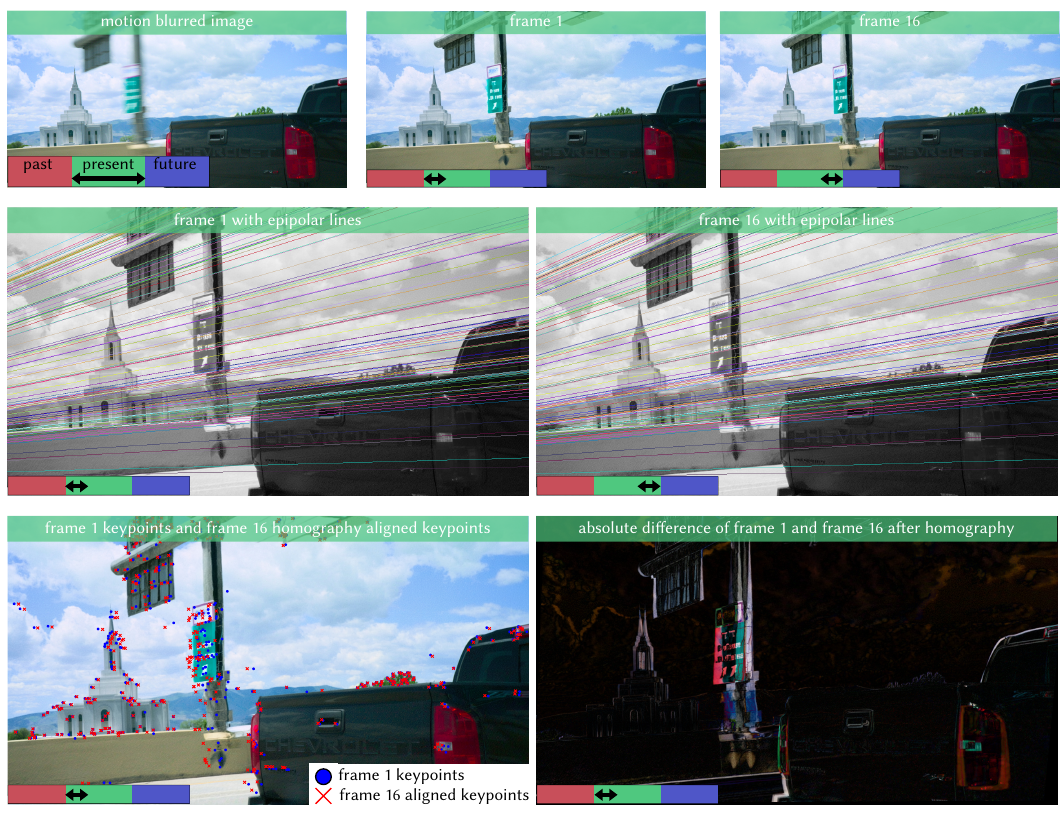}
  \caption{We assess 3D consistency of an image with motion blur due to camera movement. We visualize the epipolar lines, the residual movement of keypoints after applying a 2D homography that best aligns the two frames, and the absolute difference of the first frame with the homography-warped version of the last frame. We find the epipolar lines to be consistent with the forward (left to right) camera motion. Additionally, we observe parallax between the traffic sign and the background landscape; as a result, the homography cannot accurately model the traffic sign. }
  \label{fig:geometric_consistency_1}
\end{figure*}

\subsection{Consistency}
\label{sec:consistency}

We provide further assessment of the model based on (1) the overall 3D consistency of generated videos and (2) the consistency of the output frames with the input motion-blurred image.

% We also aimed to test the consistency of our generated videos in two ways.
% First, we averaged the generated frames and compared them to the input blurred image.

\begin{figure*}[]
    \includegraphics[width=1.00\textwidth]{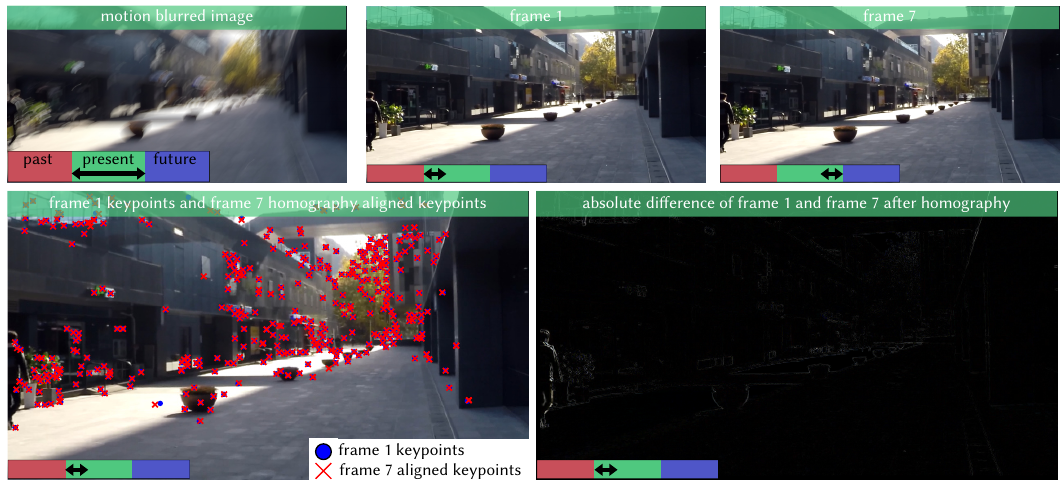}
  \caption{We assess geometric consistency of an image from the GoPro dataset~\cite{nah2017deep} with motion blur from a panning camera. We visualize the movement of keypoints after applying the 2D homography that best aligns them, and the absolute difference of the first frame with the homography-warped version of the last frame. In this case, apparent motion in the video can be accurately modeled with a homography as the viewpoint did not change. Note that in this case our approach reveals the presence of independent scene motion in the blurry photo (person's outline in the difference image). }
  \label{fig:geometric_consistency_3}
\end{figure*}

\paragraph{3D consistency}We evaluate the 3D consistency of the model output by analyzing generated videos from scenes with motion blur due to viewpoint movement. 
We apply SIFT~\cite{sift} and RANSAC~\cite{ransac} to the first and last generated frames to detect feature correspondences and compute the fundamental matrix~\cite{hartley2003multiple}. Additionally, we utilize RANSAC~\cite{ransac} to find the 2D homography that best explains the keypoint correspondences between both images.  
We visualize the epipolar lines and the homography applied to both the keypoints and images in Figures~\ref{fig:geometric_consistency_1}–\ref{fig:geometric_consistency_3}. We find in the scene with the truck (Figure~\ref{fig:geometric_consistency_1}), the epipolar lines are consistent with the forward camera motion. Additionally, we observe that while the homography fits the background, it produces large errors on the traffic sign seen in the absolute difference map, highlighting parallax caused by the scene’s 3D structure. We observe a similar effect in Supp. Figure S1, where the foreground bushes closer to the camera exhibit noticeable parallax. Finally, in Figure~\ref{fig:geometric_consistency_3}, we observe that for a scene with a panning camera (i.e., no viewpoint change), the motion is accurately modeled via a 2D homography as expected. 
These visualizations appear subjectively consistent with the scene geometry, suggesting that our model produces geometrically-plausible outputs with a reasonable degree of 3D consistency, even in the absence of explicit 3D supervision. 

\begin{figure*}[]
  \includegraphics[width=\textwidth]{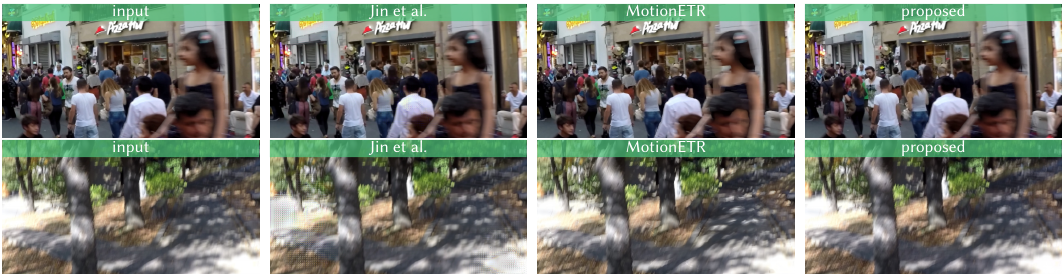}
  \caption{Comparison of input blurred image and the averaged frames of videos generated by Jin et al.~\shortcite{jin2018learning}, MotionETR~\shortcite{zhang2021exposure}, and our proposed method, from the GoPro dataset~\cite{nah2017deep}. }
  \label{fig:motionblurconsistency}
\end{figure*}

\paragraph{Consistency with input motion-blurred image.} We compare the motion-blurred image to the image created by averaging together the generated video frames.
Table~\ref{tab:motionblurconsistency} reports quantitative results of this comparison using PSNR, and Figure~\ref{fig:motionblurconsistency} provides qualitative visualizations. We note that this metric is not informative on its own because the trivial solution---repeatedly outputting the input motion-blurred image---achieves a perfect score.

\begin{table}[t]
    \caption{Motion blur consistency results of our method and baselines on the GoPro and BAIST++ datasets. We report PSNR values (dB) between the input motion-blurred image and the average of generated frames.}
    \small
    \begin{tabular}{lcc}
        \toprule
        Dataset & Method & PSNR (dB) $\uparrow$ \\
        \midrule
        GoPro   & Jin et al.~\shortcite{jin2018learning}       & 33.64 \\
        GoPro   & MotionETR~\shortcite{zhang2021exposure}          & 32.17 \\
        GoPro   & Proposed                                     & \textbf{35.47} \\
        \midrule
        BAIST++ & Animation-from-Blur~\shortcite{zhong2022animation} & \textbf{33.76} \\
        BAIST++ & Proposed                                     & 32.32 \\
        \bottomrule
    \end{tabular}
    \label{tab:motionblurconsistency}
\end{table}

We note that other methods, such as diffusion posterior sampling (DPS)~\cite{chung2023diffusion}, can explicitly enforce consistency with the motion-blurred image in their objective function. In contrast, our method achieves consistency as a natural result of our fine-tuning process on motion-blurred images synthesized from videos, i.e. the model is fine-tuned to generate frames that reproduce the input motion-blurred image when averaged. We attempted applying PSLD~\cite{psld2023}, a DPS-based method designed for latent space models, to enforce consistency for our $1280\times720$ generated videos. However, we found it would require a prohibitive 380 GB of memory to backpropagate through the variational autoencoder (VAE) due to its high parameter count.

\begin{comment}
To evaluate 3D consistency, we investigate model performance on scenes where motion blur is caused by camera movement.
We consider a few such scenes (shown in Supp.\ Fig.~S5-S7) and perform keypoint matching across the first and last generated video frames using SIFT~\cite{sift}.
Then, we estimate the fundamental matrix and visualize the epipolar geometry. We also apply homography-based warping to assess parallax effects and to show that, for scenes where motion blur is due to a panning camera, the generated motion is accurately modeled via homography. Our analysis shows the generated videos contain realistic camera geometry and parallax and can convey 3D-consistent structure. See Supp.\ Section 1.4 for more details.
\end{comment}

% This procedure results in disparity values that plausibly align with scene geometry, suggesting that the generated videos can convey 3D-consistent structure. See Supp.\ Section 1.4 for more details.}
% consistency over time (enabling correspondence matching) and in terms of 3D structure. 
% On generated videos from these scenes, we can perform keypoint matching, compute fundamental matrices, and rectification.
% In these rectified images, we find that the disparities of keypoints indicate plausible scene geometry in Supp. Section ????. }

\subsection{Exposure Interval Control}
\label{sec:exposure_interval_control}

Finally, we evaluate how well our model controls the exposure interval of each generated video frame using the exposure interval embedding introduced in Section~\ref{sec:architecture}.
We use a subset of 20 simulated motion-blurred images from the GoPro test split and condition the model to output video frames with varying exposure durations and start/end times. Specifically, we consider two general configurations: one with consecutive exposure intervals that divide the input image's exposure evenly into 2, 4, 8, or 16 frames, and another that includes a ``dead time'' between adjacent frames.
%
% The output frames achieve a high level of consistency with the ground truth in terms of image quality metrics and FVD (see Table~\ref{tab:exposure_control}). 
% We also find that, qualitatively, the amount of motion blur in the generated video frames increases when the exposure duration is increased, as expected.
% Finally, we find that our method outperforms an alternative version of the architecture that uses a different exposure interval encoding scheme, as seen in Table~\ref{tab:alternative_exposure_control}. Additional qualitative results showcasing this controllability are visualized in the supplemental webpage. }

For the first case, we simulate motion-blurred input images by averaging 16 consecutive frames captured at 1920 FPS from 20 videos in the GoPro test split, following the procedure described in Section~\ref{sec:implementation_details}. The ground-truth outputs correspond to different uniform subdivisions of this 16-frame exposure window, with our model conditioned on the corresponding exposure interval encodings. Concretely, the input is a motion-blurred image formed by averaging 16 ground-truth frames $[f_1\dots f_{16}]$, and we generate outputs at varying temporal resolutions:

\begin{itemize}

    \item 2 frames (8$\times$ duration): [f$_1\dots$f$_8$], [f$_9\dots$f$_{16}$],  
    \item 4 frames (4$\times$ duration): [f$_1\dots$f$_4$], [f$_5\dots$f$_8$], [f$_9\dots$f$_{12}$], [f$_{13}\dots$f$_{16}$],  
    \item 8 frames (2$\times$ duration): [f$_1$, f$_2$], [f$_3$, f$_4$], \dots, [f$_{15}$, f$_{16}$], 
    \item 16 frames (1$\times$ duration): [f$_1$], 
    [f$_2$], \dots, [f$_{16}$],
\end{itemize}

\noindent where $([f_s\dots f_e])$ is the average of frames $f_s,\dots,f_e$. This setup evaluates how well our method adapts to different frame durations under the same blurred input.

For the second case, we introduce dead time between frames. Here, we average 32 consecutive 1920 FPS ground-truth frames $[f_1\dots f_{32}]$ to form the input motion-blurred image and then generate 16 output frames, each corresponding to disjoint exposure intervals with a one-frame gap. Specifically, the outputs are [f$_1$], [f$_3$], [f$_5$], \dots, [f$_{31}$]. This setting tests whether our model performs well when exposure intervals are non-consecutive.

Our model maintains strong performance, demonstrating robustness to both longer exposure durations with stronger blur and to disjoint exposure intervals. Specifically, Table~\ref{tab:exposure_control} and qualitative comparisons in our supplemental webpage demonstrate that our approach provides fine-grained control over exposure intervals and remains effective under various sampling conditions.
As the number of output frames increases (i.e., shorter exposures), our model successfully reconstructs high-quality outputs, with only modest degradation in PSNR, SSIM, and perceptual metrics. When the exposure intervals are separated with dead time, performance decreases due to the increased ambiguity and blur size. Nevertheless, the model still generates plausible videos, showing it generalizes to non-consecutive start/end times for the generated frames.

% \begin{figure*}[h!]
%   \includegraphics[width=\textwidth]{figures/intervalcontrol1.pdf}
%   \caption{The exposure interval encoding enables video generation with different exposure intervals for each frame by specifying the start and end times of each frame. Given a 16-frame motion-blurred image, we visualize the model predictions in four modes: 2 frames (8$\times$ duration), 4 frames (4$\times$ duration), 8 frames (2$\times$ duration), 16 frames (1$\times$ duration). As the duration of predicted frames decreases (from the first row to the fourth row), the amount of motion blur is reduced. Finally, given a 32-frame motion-blurred image, the model predicts 16 sharp frames, each separated by one frame of dead time, which further demonstrates the controllability of the approach.}
%   \label{fig:exposure_control}
% \end{figure*}

\paragraph{Alternative exposure control scheme.}
We also compare against an alternative exposure interval encoding scheme.
Instead of explicitly providing the per-frame exposure intervals as input, we fine-tuned a new model whose temporal conditioning signal consists of (1)~the start time of the first output frame, (2)~the end time of the last output frame, (3)~the (uniform) duration of individual frames, and (4)~the number of frames.
This scheme contains equivalent information but encodes the intervals implicitly.
All other details regarding the model---sinusoidal encoding, linear layer, and addition to latent patches---are kept the same.
We repeat the two evaluations discussed above for this model as well.
Comparing Tables~\ref{tab:exposure_control} and \ref{tab:alternative_exposure_control}, it is clear that our choice of per-frame exposure interval encoding is superior to this alternative encoding scheme.
Please refer to the supplemental webpage for additional qualitative comparisons.

\begin{table}[t]
    \caption{Evaluation of exposure interval control on the GoPro Dataset. Results are reported using bidirectional patch PSNR, SSIM, and LPIPS, along with FVD.}
    \small
    \resizebox{\columnwidth}{!}{%
    \begin{tabular}{lccccc}
        \toprule
        Setting  & PSNR$_p$ $\uparrow$ & SSIM$_p$ $\uparrow$ & LPIPS$_p$ $\downarrow$ & FVD $\downarrow$ \\
        \midrule
        2 frames / 8$\times$ duration & 31.10 & 0.92 & 0.014 & 156.31 \\
        4 frames / 4$\times$ duration  & 27.65 & 0.86 & 0.024 & 205.89 \\
        8 frames / 2$\times$ duration  & 27.07 & 0.84 & 0.027 & 154.53 \\
        16 frames / 1$\times$ duration & 26.15 & 0.82 & 0.028 & 124.73 \\
        \midrule
        16 frames w/ 1 frame dead time & 23.00 & 0.72 & 0.054 & 206.74 \\
        \bottomrule
    \end{tabular}}
    \label{tab:exposure_control}
\end{table}

\begin{table}[t]

    \caption{Evaluation of the alternate method for exposure interval control on the GoPro Dataset. Results are reported using bidirectional patch PSNR, SSIM, and LPIPS, along with FVD.}
    \small
    \resizebox{\columnwidth}{!}{%
    \begin{tabular}{lcccc}
        \toprule
        Setting  & PSNR$_p$ $\uparrow$ & SSIM$_p$ $\uparrow$ & LPIPS$_p$ $\downarrow$ & FVD $\downarrow$ \\
        \midrule
        2 frames / 8$\times$ duration  & 25.25 & 0.82 & 0.041 & 353.60 \\
        4 frames / 4$\times$ duration  & 24.25 & 0.78 & 0.042 & 309.08 \\
        8 frames / 2$\times$ duration  & 24.39 & 0.77 & 0.045 & 353.24 \\
        16 frames / 1$\times$ duration & 24.52 & 0.78 & 0.046 & 274.96 \\
        \midrule
        16 frames w/ 1 frame dead time & 20.23 & 0.63 & 0.100 & 433.62 \\
        \bottomrule
    \end{tabular}}
    \label{tab:alternative_exposure_control}
\end{table}

\begin{table}[t]
    \caption{Evaluation of our exposure interval embedding without the sinusoidal projection. Results are reported using bidirectional patch PSNR, SSIM, and LPIPS, along with FVD.}
    \small
    \resizebox{\columnwidth}{!}{%
    \begin{tabular}{lcccc}
        \toprule
        Setting  & PSNR$_p$ $\uparrow$ & SSIM$_p$ $\uparrow$ & LPIPS$_p$ $\downarrow$ & FVD $\downarrow$ \\
        \midrule
        2 frames / 8$\times$ duration  & 28.27 & 0.88 & 0.026 & 304.79 \\
        4 frames / 4$\times$ duration  & 26.21 & 0.82 & 0.037 & 329.05 \\
        8 frames / 2$\times$ duration  & 25.21 & 0.78 & 0.042 & 332.27 \\
        16 frames / 1$\times$ duration & 24.35 & 0.75 & 0.046 & 325.27 \\
        \midrule
        16 frames w/ 1 frame dead time & 20.93 & 0.63 & 0.081 & 577.26 \\
        \bottomrule
    \end{tabular}}
    \label{tab:sin_ablation}
\end{table}

\paragraph{Sinuosidal embedding ablation.}

We ablate the sinusoidal projection used in our exposure interval encoding. Specifically, we train a model on the GoPro~\cite{nah2017deep} dataset under the same settings as the main paper, but replace the sinusoidal projection with a simple linear layer applied directly to the exposure intervals. Comparing Tables~\ref{tab:exposure_control} and \ref{tab:sin_ablation} suggests that the sinusoidal embedding plays a crucial role in the effectiveness of our method.

%% file: sections/conclusion.tex
\section{Discussion and Conclusion}
A surprising amount of information can be recovered from a single motion-blurred image.
Our approach offers a deeper glimpse into the moment of capture---and gestures toward the past and future existence that surrounds it. Now, we highlight a few limitations of our approach and exciting directions for future work. 

\begin{figure}
\includegraphics[width=\columnwidth]{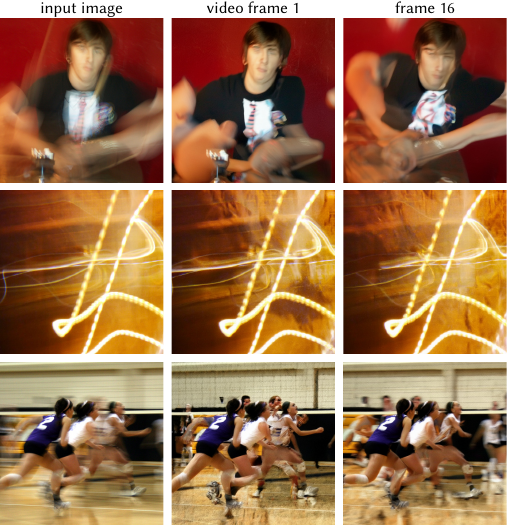}

\caption{Limitations. \textbf{(row 1)} An image of a drummer is created by compositing multiple photographs captured at different time instances. This image does not conform to our motion blur model (which assumes a single, contiguous exposure time), and so the output frames have relatively poor image quality. \textbf{(rows 2--3)} Our model fails to reconstruct videos from extremely long exposures, such as a time-lapse image of a sparkler; it can also produce artifacts when applied to images with significant motion blur due to a fast-panning camera and scene motion, such as for this sports photo. \textit{Photos from Flickr: (top) \textcopyright{} Andrew Toskin, CC BY-SA 4.0; (middle) \textcopyright{} Matt Page, CC BY-SA 2.0; (bottom) \textcopyright{} Annie Chartrand, used with permission.}}%  \textit{Photos: }}
\label{fig:limitations}

\end{figure}

As shown in Figure~\ref{fig:limitations} (row 1), the model struggles to recover videos from images that deviate from our assumed model of motion blur---such as photos created by compositing images together that were captured with separated exposure times.
Similarly, scenes with extreme motion blur, including time-lapse images and images with a combination of fast camera panning and complex scene motion, often fail to generate plausible videos  (Figure~\ref{fig:limitations}, rows 2--3).
These failure cases could potentially be addressed by fine-tuning the model on more diverse training data, including examples with time-lapse effects or other types of motion blur.

Our work has only scratched the surface of how large, pre-trained video diffusion models might be used to recover scene information from image degradations.
For instance, defocus blur and chromatic aberration may serve as additional cues for inferring scene geometry, and other optical effects could be similarly exploited.
Moreover, our findings raise the intriguing possibility of end-to-end design~\cite{sitzmann2018end,tseng2021differentiable} of optical degradations in tandem with large pre-trained models to purposefully encode information into an image for a downstream task (such as video generation).
With ongoing advances in generative modeling and access to large-scale video datasets, we anticipate many new opportunities at the intersection of image or video restoration and video generation.

%% file: sections/acknowledgements.tex
\begin{acks}
We are grateful to Konstantinos Derpanis for encouraging the first author at an early stage of this project. Thank you to Annie Chartrand for the image of the volleyball players. DBL and KNK, and MSB acknowledge support of NSERC under the RGPIN program. DBL also acknowledges support from the Canada Foundation for Innovation and the Ontario Research Fund. This study was funded in part by the Canada Research Chair program.
\end{acks}